\pdfoutput=1

\documentclass[11pt]{article}

\usepackage[preprint]{acl}

\usepackage{times}
\usepackage{latexsym}

\usepackage[T1]{fontenc}

\usepackage[utf8]{inputenc}

\usepackage{microtype}

\usepackage{inconsolata}

\usepackage{graphicx}

\usepackage{multirow}
\usepackage{amsmath}
\usepackage{amssymb}
\usepackage{algorithm}
\usepackage{algorithmic}

\def\RR{\mathbb{R}}
\def\NN{\mathbb{N}}

\usepackage[most]{tcolorbox}
\newcommand{\specialcell}[2][c]{%
  \begin{tabular}[#1]{@{}c@{}}#2\end{tabular}}

\title{Draft on the Fly:\\Adaptive Self-Speculative Decoding using Cosine Similarity}

\author{
  \textbf{Michael R. Metel\textsuperscript{1}},
  \textbf{Peng Lu\textsuperscript{2}},  
  \textbf{Boxing Chen\textsuperscript{1}},
  \textbf{Mehdi Rezagholizadeh\textsuperscript{1}}, \textbf{Ivan Kobyzev\textsuperscript{1}}
\\
  \textsuperscript{1}Huawei Noah’s Ark Lab,
  \textsuperscript{2}DIRO, Universit\'e de Montr\'eal, Canada  
\\
  \small{
    \textbf{Correspondence:} \href{mailto:michael.metel@h-partners.com}{michael.metel@h-partners.com}
  }
}

\begin{document}
\maketitle
\begin{abstract}
We present a simple on the fly method for faster inference of large language models. Unlike other (self-)speculative decoding techniques, our method does not require fine-tuning or black-box optimization to generate a fixed draft model, relying instead on simple rules to generate varying draft models adapted to the input context. We show empirically that our light-weight algorithm is competitive with the current SOTA for self-speculative decoding, while being a truly plug-and-play method. 
\end{abstract}

\section{Introduction}

The main bottlenecks of transformer-based large language model (LLM) inference are the sequential generation of tokens, due to its autoregressive nature, and the need to reload the KV cache, causing inference to be memory bandwidth bound \cite{shazeer2019}. Both of these bottlenecks prevent the full use of computing resources. Speculative decoding is a method to decrease the inference time of LLMs by taking advantage of this available compute capacity through the observation that the time of processing a single input token is roughly the same as processing a small number of tokens in parallel. 

After processing the input context and generating the first token $t_1$ using a model $M$, instead of sequentially generating tokens until completion, speculative decoding uses a smaller {\it draft} model $D$ to generate a short sequence of tokens $[\hat{t}_2,\hat{t}_3,...]$. The tokens $[t_1,\hat{t}_2,\hat{t}_{3},...]$ are then processed in parallel by $M$ for verification. The first rejected token $\hat{t}_r$ is replaced by $M$'s predicted token $t_r$. The process is then repeated, generating a new sequence of tokens $[\hat{t}_{r+1},\hat{t}_{r+2},...]$ for verification. 

When using greedy decoding, the verification process is to simply ensure that $\hat{t}_i=t_i$ for each drafted token $\hat{t}_i$. When sampling is used, a clever method has been devised to accept tokens which ensures that the distribution they were sampled from matches that of $M$ \cite{chen2023,leviathan2023}. Simply put, speculative decoding has the ability to speed up inference without sacrificing accuracy. 

The ability to speed up inference also stems from the varying difficulty in predicting the next token over the generated sequence, 
where even quite small draft models are capable of predicting a percentage of tokens correctly. Choosing a $D$ to maximize speedup then requires balancing its speed of token generation with the accuracy of the tokens it generates.   

In this work we propose a self-speculative decoding algorithm, where $D$ is chosen as a subnetwork of $M$, generated on the fly and adapted to the input context. Compared to the current state of speculative decoding described in Section \ref{related}, our method is simple to implement as a truly plug-and-play method, avoiding the challenges of finding a fixed draft model using fine-tuning or black-box optimization. In Section \ref{exp} we observe that despite the simplicity of our proposed Adaptive Draft Model Generator (ADMG, Algorithm \ref{alg1}), our method is competitive with the current SOTA of self-speculative decoding. 

\section{Related Work}
\label{related}

Speculative decoding was first developed in \cite{chen2023,leviathan2023}. These works proposed using a second smaller model to draft tokens from. Acquiring such a draft model is in general challenging, and will typically require fine-tuning \cite{cai2024, li2024, zhang2023}. Sufficient memory must also be available to store a second model as well as save both models' KV caches. 

To alleviate the need for a separate draft model, self-speculative decoding was developed in Draft \& Verify \cite{zhang2023}, where $D$ is chosen as a subnetwork of $M$ by removing attention and MLP layers. Choosing which layers to remove to minimize the average inference time is a binary optimization problem (NP-hard) of a black-box objective function. An approximate solution to this problem is found using Bayesian optimization with a small number of training samples. This is a time consuming process (see Section \ref{exp}), with the solution depending on the model architecture, dataset, and computing environment.

Kangaroo \cite{liu2024} can be seen as a hybrid approach between using a fine-tuned draft model and self-speculative decoding, where $D$ consists of the first $l\in\NN$ layers of $M$, a fine-tuned adapter layer, and the LM Head of $M$.

Medusa \cite{cai2024} is an adjacent method to speculative decoding which uses multiple decoding heads to predict tokens in parallel, avoiding the challenges of obtaining an appropriate draft model. In its simplest form, these additional heads must be trained with the original weights of $M$ kept frozen. 

EAGLE \cite{li2024} also does not use a separate draft model, but drafts tokens using an additional decoder layer which must be trained. The added decoder layer takes as input the embedding of the last generated token and the input to the LM head of the penultimate generated token, with its output fed into the LM head of $M$ to predict the next token.

Compared to these related works, a main benefit of our method is that it does not require the use of any fine-tuning or black-box optimization. Fine-tuning can be challenging when the training set of $M$ is not available, as this can result in a shift in the output distribution of $D$ compared to $M$, making the token verification stage of speculative decoding more challenging \cite{cai2024}. Similarly, for Draft \& Verify, the chosen draft model can be sensitive to the calibration data used, and its performance may suffer if an appropriate dataset is not available, see Section \ref{dis} for further discussion.  

\section{Adaptive Self-Speculative Decoding}

In order to generate draft models on the fly, our main technique is to use a simple thresholding rule to remove layers from the original model based on the cosine similarities of the hidden states of the input context.

\subsection{Removing Attention Layers Based on Cosine Similarity}

Our proposed Adaptive Self-Speculative Decoding (ASD) method uses the cosine similarity of the hidden states before and after each attention layer as an estimate of its importance, based on the intuition that the closer the cosine similarity is to 1 for a given attention layer, the smaller the effect of removing it should be on the model's accuracy. Our decision to only remove attention layers is motivated by observations from computing the average cosine similarity (ACS) of every attention and MLP layer, and the average compute time of attention and MLP layers of Llama-2-13B over the input context of 1000 samples of the training set of CNN/DM, summarized by the statistics given in Table \ref{T1}. 
\begin{table}
\centering
	\begin{tabular}{lll}
		\hline
		& Attention & MLP \\
            \hline
		Minimum ACS & 0.581  & 0.854  \\
		Maximum ACS & 0.998 & 0.991  \\
		Mean ACS	& 0.953 & 0.942 \\
		Median ACS & 0.977 & 0.960 \\
		Mean Time (ms) & 1.339  & 0.768 \\
		\hline		
	\end{tabular}
\caption{Statistics of the average cosine similarity (ACS) and the average compute time of the layers of Llama-2-13B based on the input context of 1000 training set samples of CNN/DM.}
\label{T1}
\end{table}
The following three observations motivated the focus on only removing attention layers based on cosine similarity: 
\begin{enumerate}
\item The higher mean and median ACS of attention layers indicate that on average more attention layers than MLP layers can be removed.
\item The range of the ACS is larger for attention layers, considering Maximum ACS - Minimum ACS, allowing for an easier differentiation of removable attention layers.
\item The average attention compute time is $1.74\times$ greater than that of MLP layers.
\end{enumerate}

Considering the added compute time to inference from having to calculate the cosine similarity and determine which layers to remove, we concluded that focusing solely on attention layers would give the best return in terms of inference speedup, given that more attention layers will be removable, it will be easier to determine which layers to remove, and a greater reduction in draft time will be achieved per removed layer.

\subsection{Implementation \& Further Heuristics}

When performing inference, the input context serves as a natural calibration set to adaptively choose which attention layers to remove from $M$ to form $D$ for the given instance. 

The first step of inference is to pass the entire input context through the original model $M$ in parallel to generate the first token and populate the KV cache. As the hidden states of the input pass through each layer, ASD computes the ACS of each attention layer over the input sequence length in order to choose $D$ for the following token generation stage. 

In particular, let $X^l\in\RR^{S\times H}$ and $Y^l\in\RR^{S\times H}$ be the hidden states before and after the $l^{\text{th}}$ attention layer for $l=1,...,L$, where $S$ is the input sequence length, $H$ is the hidden size, $L$ is the number of layers, and assuming a batch size of 1. We first compute the ACS over the input sequence, $C_l:=\frac{1}{S}\sum_{s=1}^{S}\frac{\langle X_s^l,Y_s^l\rangle}{||X_s^l||_2||Y_s^l||_2}$. For a fixed constant $\alpha\in(0,1)$, all attention layers with $C_l\geq \alpha$ are removed from $M$ to generate $D$ for the current generation task, which will be refered to as {\it CS thresholding}.

Even though this method enabled significant inference speedup on its own, in order to further speed up inference, we found two simple deterministic rules to be effective:
\begin{enumerate}
\item Remove every $m^{th}$ attention and MLP layer.
\item Do not remove the last $n$ attention and MLP layers when using rule 1 or CS thresholding.
\end{enumerate}

An insightful hypothesis for the effectiveness of rule 1 was proposed in \cite{sajjad2020}: In an $M$ with redundancy in its layers, neighbouring layers contain similar information, hence if layer $l$ is removed, its information is still largely contained in layers $l-1$ and $l+1$. 

Rule 2 was initially proposed to preserve the first and last $n$ layers, but by choosing $n<m$ in our implementation (Table \ref{T2}), and from low observed ACS in the first $n$ layers, this extra condition became unnecessary. The analysis in \cite{sun2024} gives support for this rule, as it was found that middle layers, matching closely to layers $\{l : n< l \leq L-n\}$ for our given $n$ in Table \ref{T2}, share the same representation space, allowing the output from layer $k$ to be interpretable by layer $l$, for $l>k$. We also note that the importance of the last few layers has been previously observed in \cite{gromov2024}.
   
Our Adaptive Draft Model Generator (ADMG) is presented as Algorithm \ref{alg1}, which outputs the attention and MLP layers to remove. An ablation study on ADMG is presented in Section \ref{dis}.
\begin{algorithm}
	\caption{Adaptive Draft Model Generator (ADMG)} 
	\begin{algorithmic}        
	\STATE {\bfseries Input:} $C\in[-1,1]^{L}$; $\alpha\in(0,1)$; $m,n\in\NN$	
            \STATE $\text{remove}_{\text{ATTN}}=\{l : C_l\geq \alpha, l\leq L-n\}$
		  \STATE $\text{remove}_{\text{MLP}}=\{l : l=jm, m\in\NN, l\leq L-n\}$
            \STATE $\text{remove}_{\text{ATTN}}=\text{remove}_{\text{ATTN}}\cup\text{remove}_{\text{MLP}}$    
            \STATE {\bfseries Output:} $\text{remove}_{\text{ATTN}}$, $\text{remove}_{\text{MLP}}$  
	\end{algorithmic}
	\label{alg1}
\end{algorithm}

\section{Experiments}
\label{exp}

We compare our method to the self-speculative decoding algorithm of Draft \& Verify, which is most similar to our work. We conducted the same (fine-tuned) Llama-2-13B experiments found in the body of their paper without changing any of the chosen hyperparameters. 

The experiments cover three models and three datasets: Llama-2-13B and Llama-2-13B-Chat \cite{touvron2023} evaluated on 1000 random test set samples of CNN/DM \cite{nallapati2016} and XSUM \cite{narayan2018}, and CodeLlama-13B \cite{roziere2023} evaluated on HumanEval \cite{chen2021}. CNN/DM and XSUM are both summarization tasks whereas HumanEval is a Python code generation task. The test sets were sampled using the same random seed as in Draft \& Verify's experiments. 

An important factor which determines the level of inference speedup is the number of tokens drafted before verifying their accuracy with $M$. We used the Adaptive Draft-Exiting Mechanism proposed in Draft \& Verify \cite[Section 3.4]{zhang2023}, keeping the hyperparameters equal to their tuned values \cite[Table 6]{zhang2023} to match their experiments.

We conducted all experiments using two V100-32GB GPUs. Draft \& Verify relies on Bayesian optimization to find their draft model, which is sensitive to the computing environment. In order to get the best performance from their method, we reran their optimization code to generate draft models tuned to our environment. In total three draft models were generated, one for each model,  which took on average 11 hours to generate per model. 

For our ADMG (Algorithm \ref{alg1}), we used the hyperparameters in Table \ref{T2}. The only difference over the experiments was that higher speedup was observed by incrementing $m$ and $n$ by 1 for the CodeLlama-13B experiments. Since CodeLlama-13B is fine-tuned for a more specific task compared to the other models, we believe that there may be less redundant layers which can be easily removed using rules 1 and 2.
\begin{table}
\centering
	\begin{tabular}{lccc}		
		\hline
         Model &  $\alpha$ & $m$ & $n$ \\
            \hline
		 Llama-2-13B & 0.985 & 3 & 2\\
		 Llama-2-13B-Chat & 0.985 & 3 & 2\\
		 CodeLlama-13B & 0.985 & 4 & 3\\			
        \hline				          
	\end{tabular}        
	\caption{Hyperparameters used for ADMG (Alg. \ref{alg1})}
 \label{T2}
\end{table}

The results of the experiments are presented in Tables \ref{T3} and \ref{T4}. We observe that both methods have similar performance overall, with Draft \& Verify performing better on the Llama-2-13B experiments, ASD having better performance on the CodeLlama-13B experiments, with mixed performance observed for Llama-2-13B-Chat. 
\begin{table}
\centering

\resizebox{\columnwidth}{!}{%
    \begin{tabular}{lllcccc}
    \hline
    \multirow{1}{*}{Model} & \multirow{1}{*}{T} & \multirow{1}{*}{Method} & \multirow{1}{*}{CNN/DM} & \multirow{1}{*}{XSum} \\
    \hline
    \multirow{4}{*}{\specialcell{Llama-2\\-13B}} & \multirow{2}{*}{0.0}  & D\&V & {\bf 1.495}\small{$\times$} & {\bf 1.415}\small{$\times$} \\
    &  & ASD & 1.443\small{$\times$} & 1.383\small{$\times$} \\
    \cline{2-5}
    & \multirow{2}{*}{0.2} & D\&V & {\bf 1.479}\small{$\times$} &  {\bf 1.383}\small{$\times$} \\
    & &	ASD & 1.422\small{$\times$} & 1.350\small{$\times$} \\
    \hline		
    \multirow{4}{*}{\specialcell{Llama-2\\-13B-Chat}} & \multirow{2}{*}{0.0} & D\&V & 1.238\small{$\times$} & {\bf 1.143}\small{$\times$}\\
    &  & ASD & {\bf 1.247}\small{$\times$} & 1.110\small{$\times$}\\
    \cline{2-5}
    & \multirow{2}{*}{0.2} & D\&V & 1.219\small{$\times$} &  {\bf 1.134}\small{$\times$}\\
    &  & ASD & {\bf 1.238}\small{$\times$} & 1.102\small{$\times$} \\
    \hline
    \end{tabular}
 }
\caption{Inference speedup of Draft \& Verify (D\&V) and Adaptive Self-Speculative Decoding (ASD) methods compared to vanilla autoregressive generation using different temperatures (T).}
\label{T3}
\end{table}

\begin{table}
\centering

	\begin{tabular}{cccc}		
		\hline
         Model &  T & Method & Speedup \\
            \hline
		\multirow{4}{*}{CodeLlama-13B} &  \multirow{2}{*}{0.0} & D\&V &  1.282\small{$\times$} \\
		& & ASD  &  {\bf 1.365}\small{$\times$} \\	\cline{2-4}	
		& \multirow{2}{*}{0.6} & D\&V &  1.282\small{$\times$} \\
		& & ASD  &  {\bf 1.340}\small{$\times$} \\			
        \hline				  
	\end{tabular}
	\caption{HumanEval inference speedup of Draft \& Verify (D\&V) and Adaptive Self-Speculative Decoding (ASD) compared to vanilla autoregressive generation using different temperatures (T).}
    \label{T4}
\end{table}

\subsection{Discussion \& Ablation Study}
\label{dis}

It may be surprising that choosing a draft model by simple rules can give on par performance with a costly optimization method, but there are no guarantees on the solution quality when using normal Bayesian optimization to solve a discrete optimization problem by rounding the suggested continuous points, as was done in Draft \& Verify. This may result in the algorithm getting stuck by revisiting previously sampled points \cite{luong2019}, which was in fact observed when running their implementation.

The largest difference in inference speedup between Draft \& Verify and ASD is observed in the CodeLlama-13B experiments. Given that HumanEval does not have a training set, Python samples from the StarCoder \cite{li2023} training dataset were used to generate Draft \& Verify's $D$, as was done in their experiments. We believe the high relative speedup of ASD demonstrates the robustness of ADMG, and the potential sensitivity to differences in the distribution of the calibration and test datasets for their method. The implication of this being that in general, it may be challenging for Draft \& Verify to remain on par with our method when it cannot be guaranteed that the difference between the calibration data and the data it processes through time will remain small.

We end this section by giving an ablation study on ADMG, showing the inference speedup achieved when generating draft models using subsets of its 3 rules, presented in Table \ref{T5}. We observe that the majority of the inference speedup comes from CS thresholding, with increased inference speed when adding rule 2, and then again when adding rule 1 (ADMG). We also observe that even without using CS thresholding, inference speedup is achieved only using rules 1 and 2.    
\begin{table}
\centering
    \begin{tabular}{lc}		
    \hline
    Drafting Generation Rules &  Speedup\\
    \hline
    CS thresholding & 1.307\small{$\times$} \\
    CS thresholding \& rule 2 & 1.340\small{$\times$} \\	
    Rules 1 \& 2 & 1.179\small{$\times$} \\    		
    ADMG & 1.443\small{$\times$} \\
    \hline				          
    \end{tabular}        
    \caption{Ablation study on ADMG (Algorithm \ref{alg1}) showing the inference speedup using subsets of its 3 rules for Llama-2-13B on 1000 test samples of the CNN/DM dataset using greedy decoding.}
\label{T5}
\end{table}

\section{Conclusion}
We have proposed a self-speculative decoding method to generate draft models on the fly which are adapted to the input context. Our method uses cosine similarity thresholding with simple layer removal rules to generate draft models without the need for any fine-tuning, black-box optimization, or training data. Despite our proposed method's simplicity, we found that it is competitive with the current self-speculative decoding SOTA, while being an easy to implement plug-and-play method.

\section*{Limitations}

The inference speedup achieved by ASD relies on the existence of layers which when removed, do not significantly impact the model's accuracy. This reliance on the level of redundancy in the LLM's layers will limit the potential inference speedup of ASD.  

\bibliography{custom}

\end{document}